\documentclass[letterpaper]{article} % DO NOT CHANGE THIS
\usepackage{aaai23-r2hcai}  % DO NOT CHANGE THIS
\usepackage{times}  % DO NOT CHANGE THIS
\usepackage{helvet}  % DO NOT CHANGE THIS
\usepackage{courier}  % DO NOT CHANGE THIS
\usepackage[hyphens]{url}  % DO NOT CHANGE THIS
\usepackage{graphicx} % DO NOT CHANGE THIS
\usepackage{natbib}  % DO NOT CHANGE THIS AND DO NOT ADD ANY OPTIONS TO IT
\usepackage{caption} % DO NOT CHANGE THIS AND DO NOT ADD ANY OPTIONS TO IT
\usepackage{amsmath}
\newcommand{\ind}[1]{\mathbf{1}\!\left\{#1\right\}} % indicator w/o \mathbb

\usepackage{algorithm}
\usepackage{float}   % 放在 \usepackage{algorithm} 之后均可
\usepackage{needspace}   % 提供 \needspace{<length>}
\usepackage[noend]{algpseudocode}
\floatname{algorithm}{Algorithm}
\algrenewcommand\algorithmicrequire{\textbf{Input:}}
\algrenewcommand\algorithmicensure{\textbf{Output:}}

\usepackage{placeins} % provides \FloatBarrier

\usepackage{newfloat}
\usepackage{listings}
\lstdefinelanguage{json}{
  morestring=[b]",
  morecomment=[l]{//},
  morecomment=[s]{/*}{*/},
  morekeywords={true,false,null},
  sensitive=true,
  showstringspaces=false
}
\DeclareCaptionStyle{ruled}{labelfont=normalfont,labelsep=colon,strut=off} % DO NOT CHANGE THIS
\floatstyle{ruled}
\newfloat{listing}{tb}{lst}{}
\floatname{listing}{Listing}

\usepackage{fancyhdr}
\makeatletter
\newcommand\blfootnote[1]{%
  \begingroup
  \renewcommand\thefootnote{}%
  \footnote{#1}%
  \addtocounter{footnote}{-1}%
  \endgroup
}
\makeatother

\title{EmoLoom-2B: Fast Base-Model Screening for Emotion Classification and VAD with Lexicon-Weak Supervision and KV-Off Evaluation}

\author{
    Zilin Li$^1$\equalcontrib \quad
    Weiwei Xu$^1$\equalcontrib \quad
    Xuanbo Lu$^1$ \quad
    Zheda Liu$^2$
}
\affiliations {
    \textsuperscript{\rm 1} Donghua University \quad
    \textsuperscript{\rm 2} Shanghai Jiao Tong University \\
    tzulamlee@gmail.com, yumuzaijie@gmail.com, 231310131@mail.dhu.edu.cn, zd.liu.research@gmail.com
}

\begin{document}
\maketitle

\blfootnote{\footnotesize \textit{Preprint}.}

\begin{abstract}
We present \textbf{EmoLoom-2B}, a lightweight and reproducible pipeline that turns small language models ($\sim$2B) into \emph{fast-to-screen} candidates for joint emotion classification and Valence--Arousal--Dominance (VAD). To eliminate protocol drift, we unify training and inference with a one-line JSON I/O contract and report a \emph{ParseOK} rate alongside standard task metrics. Fairness is further enforced by a public \emph{KV-off} decoding setting shared across training and evaluation. Around this backbone, we add two orthogonal regularizers: a \emph{VAD-preserving} constraint that aligns the VAD implied by the generated text with the target triplet, and a lightweight external \emph{appraisal-atom} verifier (goal attainment, controllability, certainty, fairness) that guides training without lengthening justifications. We also introduce a simple \emph{Valence Flip} augmentation and an \emph{A/B mixture} schedule with entropy-aware temperature cooling to trade off coverage and convergence during SFT. Using \emph{Qwen-1.8B-Chat} as the base model, we train on \emph{GoEmotions} and \emph{EmpatheticDialogues} and probe cross-corpus generalization on \emph{DailyDialog} with lexicon-derived weak VAD. On development sets, EmoLoom-2B attains Macro-F1 of 0.35 and VAD $(1-\mathrm{RMSE})$ of 0.94 with \emph{ParseOK} $=1.00$; a time-bounded cross-corpus quick evaluation yields Macro-F1 of 0.31 while maintaining robust validity. The recipe is budget-aware, auditable, and re-entrant, providing a dependable screening pass before heavier training or multimodal fusion.
\end{abstract}

\section{Introduction}

Understanding human emotion from language remains a central capability for socially aware AI systems, with practical impact on mental-health support, education, safety moderation, and affective conversational agents. In low-latency or resource-constrained settings, small language models (SLMs, $\sim$2B parameters) are especially attractive. Yet, despite rapid progress, the evaluation of emotion understanding---especially when \emph{classification} must co-exist with continuous \emph{Valence--Arousal--Dominance (VAD)} regression---often suffers from three persistent issues: (i) protocol drift between training and inference that turns formatting and parsing into confounders rather than signal; (ii) fairness gaps in decoding (e.g., inconsistent use of KV-cache and generation hyperparameters) that inflate or deflate scores; and (iii) weak-supervision pipelines that do not explicitly preserve VAD semantics in the generated answers, undermining consistency even when label accuracy looks strong.

\begin{figure*}[t]
  \centering
  \includegraphics[width=\textwidth]{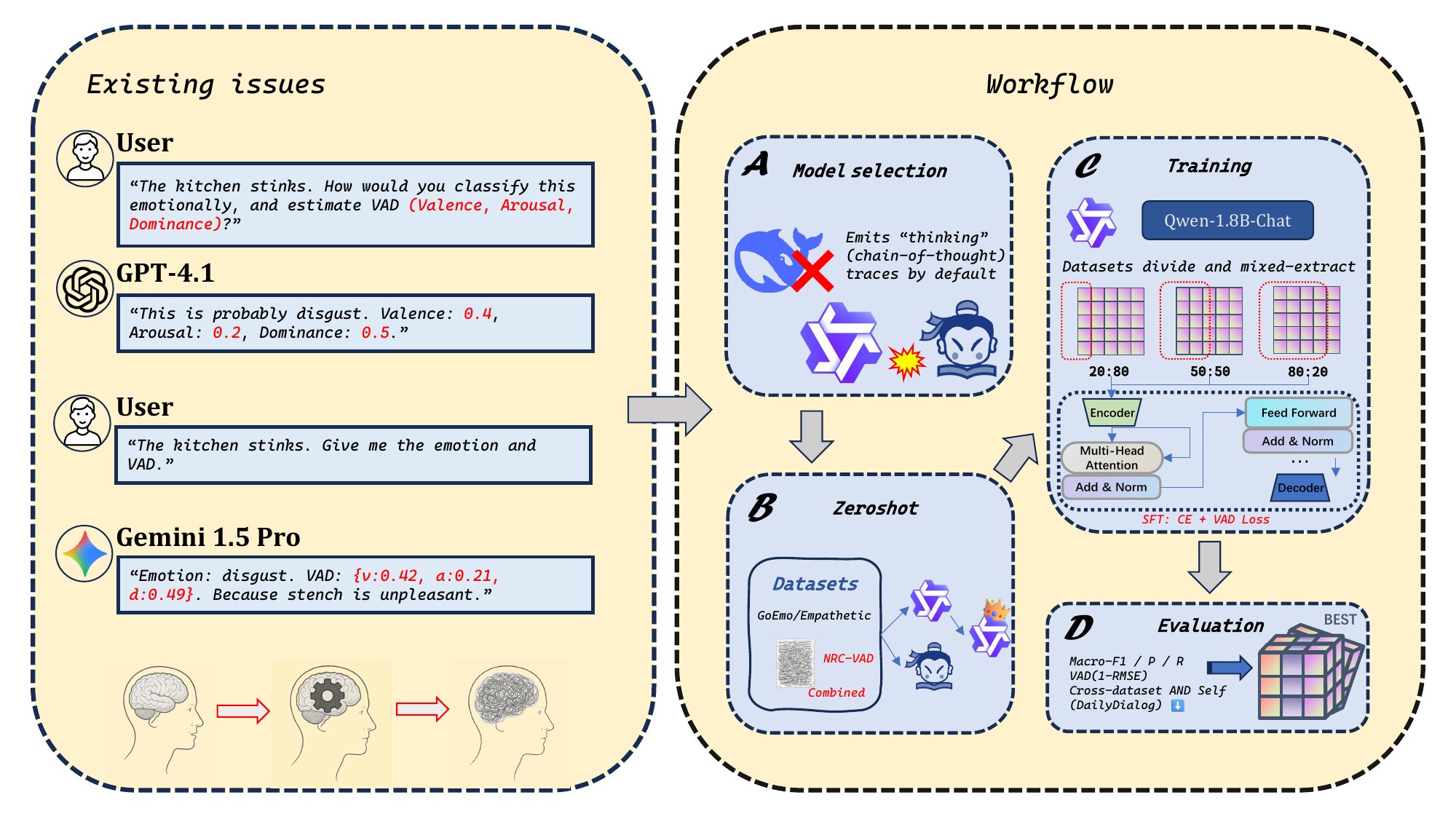}
  \caption{%
  \textbf{EmoLoom-2B overview.}
  \emph{Left: Existing issues.} When jointly asking for emotion labels and VAD, big models often drift in protocol, rely on subjective decoding, or fail to preserve VAD semantics.
  \emph{Right: Workflow (A–D).}
  \textbf{A} Model selection among open SLMs;
  \textbf{B} Zeroshot sanity checks on target datasets;
  \textbf{C} Training on Qwen-1.8B-Chat with mixed-extract splits (20:80/50:50/80:20) and CE+VAD loss under protocol-true JSON I/O;
  \textbf{D} Evaluation with Macro-F1/P/R, VAD ($1-\mathrm{RMSE}$), cross-corpus probe, and \textbf{ParseOK} validity.
  We adopt \textbf{KV-off} decoding for fairness across training and evaluation.%
  }
  \label{fig:overview}
\end{figure*}

We introduce \textbf{EmoLoom-2B}, a lightweight and reproducible pipeline that turns a broad pool of open SLMs into \emph{fast-to-screen} candidates for emotion classification and VAD, while enforcing a protocol-true and fair evaluation. The core idea is to unify data, training, and inference under a one-line JSON I/O contract and to remove avoidable evaluation variance by adopting \textbf{KV-off} decoding as a default public setting. Around this spine, we incorporate two orthogonal semantic regularizers. First, a \textbf{VAD-preserving} constraint aligns the VAD implied by the generated text with the target VAD triplet, stabilizing continuous predictions without injecting long rationales. Second, a lightweight \textbf{appraisal-atom} verifier---instantiated as a compact external classifier over goal attainment, controllability, certainty, fairness, and related “cognitive appraisal” factors---provides training-time guidance without entangling the generator with lengthy natural-language justifications. To improve polarity sensitivity, we introduce a simple \textbf{Valence Flip} augmentation that creates polarity-mirrored pairs and encourages symmetric responses. Finally, to balance coverage and convergence during SFT, we use \textbf{A/B mixture sampling with entropy-aware temperature scheduling}: high-entropy samples dominate early to widen coverage; the schedule cools over training to consolidate reliable patterns. An overview is shown in Fig.~\ref{fig:overview}.

Our practical motivation is twofold. First, many downstream deployments require a dependable \emph{screening pass} to select a base model before heavier training or multi-modal fusion. Such screening should be budget-aware, transparent, and stable across machines. Second, when emotion categories and VAD must be produced \emph{jointly}, the system should optimize for both accuracy and \emph{answer validity}: the output must be parseable, internally consistent, and semantically faithful to the requested format. EmoLoom-2B operationalizes these requirements through a minimal contract: one-line JSON with fields for multi-label \texttt{labels}, continuous \texttt{vad=\{v,a,d\}}, and a short \texttt{rationale}. We report standard metrics (Macro-F1/Precision/Recall for multi-label classification; $1-\mathrm{RMSE}$ for VAD) together with a \textbf{ParseOK} rate that quantifies adherence to the output contract. 

Concretely, we systematically screen candidate SLM backbones and adopt \emph{Qwen-1.8B-Chat} as the default base due to its favorable coverage--stability trade-off in our setup. Training uses common, steady-state practices (bf16/TF32, gradient checkpointing, cache-robust allocation) and the exact decoding configuration used for evaluation (\texttt{use\_cache=false}), eliminating train--test discrepancies. Datasets cover \emph{GoEmotions} and \emph{EmpatheticDialogues} for training and development, complemented by a cross-corpus probe on \emph{DailyDialog} with lexicon-driven weak VAD to assess generalization under domain shift. All scripts follow a single manifest and directory layout to make runs auditable and re-entrant.

\paragraph{Contributions.} 
This work offers a compact recipe for dependable, budget-aware emotion modeling with SLMs:
\begin{enumerate}
\item \textbf{Protocol-true I/O.} A one-line JSON contract unifies training and inference. We pair standard task metrics with \textbf{ParseOK} to directly measure contract adherence, reducing “format wins” that do not reflect genuine modeling gains.
\item \textbf{Fair public decoding.} We institutionalize \textbf{KV-off} decoding as the default evaluation mode and match training/inference decoding settings, removing a common source of variance and improving replicability across machines and seeds.
\item \textbf{VAD-preserving regularization.} A simple consistency loss aligns the VAD implied by the generated answer with the target VAD, improving continuous estimates without relying on lengthy, hard-to-grade rationales.
\item \textbf{Lightweight appraisal verifier.} A compact, external classifier over cognitive appraisal “atoms” provides training-time guidance, decoupling semantic checks from the generator and avoiding justification length as a hidden knob.
\item \textbf{Polarity symmetry via Valence Flip.} Polarity-mirrored pairs encourage sensitivity to valence and yield a diagnostic handle on polarity robustness at evaluation time.
\item \textbf{Coverage--convergence scheduling.} \textbf{A/B mixture sampling} with an entropy-aware temperature schedule stabilizes small-model SFT by emphasizing diverse, informative samples early and consolidating later.
\item \textbf{Fast screening under time budget.} A clear, time-bounded evaluation path enables quick comparison of candidate backbones before investing in heavier training or multi-modal integration, with unified logs, figures, and tables for paper-ready reporting.
\end{enumerate}

\paragraph{Empirical preview.}
Across development sets and a cross-corpus probe, EmoLoom-2B improves Macro-F1 and VAD error while maintaining high ParseOK; qualitatively, we see fewer format failures and steadier valence under polarity stress-tests. A 20:80 mix often gives the best F1--VAD trade-off, though dataset-dependent; detailed results and ablations follow.

\paragraph{Scope and ethics.}
We focus on text-based emotion understanding in English. Weak labels derived from public lexica are used only for training signals and diagnostics; we release aggregated metrics, not raw personal data. The pipeline is designed for reproducibility and auditability, with identical decoding for training and evaluation.

\paragraph{Paper roadmap.}
Related Work reviews prior art; Method covers the I/O contract, VAD-preserving loss, appraisal verifier, Valence Flip, and the schedule (Fig.~\ref{fig:overview}). Data and Weak-Label Generation details datasets and weak-label conversion; Implementation Details presents screening. Evaluation Protocol defines metrics; Results, Ablations and Sensitivity, and Discussion and Limitations report findings and analyze design choices; Conclusion closes.

% --- Figure: Overview at the end of Introduction ---

\section{Related Work}\label{sec:related}

\paragraph{LLM-based emotion classification and VAD.}
Early neural approaches treated emotion as single- or multi-label classification, later extending to joint prediction of discrete categories and continuous Valence--Arousal--Dominance (VAD). Recent small language models (SLMs, $\sim$2B) inherit strong text priors and can be instruction-tuned for the joint task, but evaluation often confounds modeling quality with formatting or decoding choices. Common resources include crowd-labeled dialogue and sentence datasets (e.g., GoEmotions, EmpatheticDialogues, DailyDialog) and have encouraged multi-label metrics (Macro-F1/Precision/Recall) alongside continuous VAD error. Our work follows this joint formulation but emphasizes \emph{answer validity} via a one-line JSON contract and a ParseOK rate so that improvements are not artifacts of prompt or parser idiosyncrasies \citep{dg:20,rs:19,ls:17,r:80,m:18,wkb:13}.

\paragraph{Appraisal Theory in computational emotion.}
Psychological theories describe emotions through \emph{cognitive appraisals} such as goal attainment, controllability, certainty, and fairness. Computational uses range from feature engineering and rule systems to text classifiers trained on appraisal cues. Many LLM pipelines now generate lengthy rationales to imitate human explanations, yet rationale length becomes a tuning knob and may not translate to better VAD fidelity. We instead adopt appraisal theory as a \emph{training-time constraint}: a lightweight external ``appraisal-atom'' verifier provides semantic guidance without fusing explanations into the generator, preserving simplicity, speed, and controllable failure modes \citep{occ:88,se:85,scherer:01,e:92}.

\paragraph{Lexicon-based weak supervision and VAD norms.}
Emotion and affective lexica (e.g., NRC Emotion Lexicon) and affective norm lists (e.g., Warriner’s VAD norms) have long enabled low-cost labels or priors. They are effective for coverage and quick diagnostics but are sensitive to domain shift, polysemy, and context (negation, sarcasm). In EmoLoom-2B, lexica are not treated as fixed labels; they supply weak VAD signals to (i) regularize the model through a \emph{VAD-preserving} consistency loss that aligns generated text with target VAD and (ii) support polarity stress tests via \emph{Valence Flip} augmentation. This keeps lexical priors useful while avoiding overcommitment to word-level heuristics \citep{mt:13,wkb:13,m:18,pz:06,wb:10,pl:08,rb:20}.

\paragraph{Small-model SFT and evaluation fairness.}
Instruction/SFT pipelines for small models typically vary decoding hyperparameters (temperature, top-$p$, max length) and cache usage across training and evaluation, yielding non-trivial score variance and limited replicability. KV-cache in particular changes compute and sometimes output trajectories across hardware and seeds, complicating fair comparison. We advocate a \emph{protocol-true} setup: identical JSON I/O and \textbf{KV-off} decoding for both training and evaluation, making results comparable across machines and allowing ablations to measure modeling changes rather than decoding drift \citep{bm:20,dc:19,dy:19,hbd:20,liang:22,ba:21,p:21,bgm:21}.

\paragraph{Quick, budget-aware evaluation.}
Many deployments need a rapid ``screening pass’’ to choose a base model before heavier training or multimodal fusion. Prior work on efficient benchmarking suggests time- or compute-bounded evaluation to trade breadth for speed, but such protocols are rarely standardized for emotion + VAD. EmoLoom-2B provides a simple, auditable quick-eval path—shared metrics, unified logs/figures, and ETA reporting—so candidate backbones can be compared under the same budget \citep{liang:22,rwg:20}.

\paragraph{Positioning.}
Unlike frameworks that replicate full psychological pipelines or rely on explanation-conditioned generation, we use psychology as \emph{structured constraints on the training objective}: a VAD-preserving loss and a compact appraisal verifier. Together with a protocol-true JSON contract and \textbf{KV-off} decoding, this yields a small, fast, and reproducible recipe for joint emotion/VAD modeling that is easy to audit and extend.

\section{Method}\label{sec:method}

\subsection{Task and One-line JSON I/O}\label{sec:io}
We jointly predict multi-label emotions and continuous Valence--Arousal--Dominance (VAD) per utterance $x$.
The model must return a \emph{single-line JSON}. To avoid overfull boxes in two-column layout, we typeset it with automatic line breaking:
\begin{lstlisting}[language=json,numbers=none,caption={One-line JSON I/O (wrapped for typesetting)},label={lst:json}]
{"labels":["disgust"],"vad":{"v":0.42,"a":0.21,"d":0.49},"rationale":"..."}
\end{lstlisting}
where $\mathrm{vad}\in[0,1]^3$ is rounded to two decimals and \texttt{rationale} is a short English phrase.
During evaluation we \emph{tail-scan} the generated text to find a valid JSON object; success defines \textbf{ParseOK}.  
Formally, with $N$ samples,
\[
\mathrm{ParseOK}=\frac{1}{N}\sum_{i=1}^{N}\ind{\mathrm{parse}(\hat{y}_i)}.
\]
We compute task metrics on valid outputs only and report \emph{ParseOK} side-by-side to expose formatting failures.  
Training and inference use the same prompt and decoding; KV cache is disabled for fairness (Sec.~\ref{sec:kv}).  

\paragraph{Label and VAD losses.}
Let $K$ be the number of emotion labels, $y\in\{0,1\}^K$ the multi-hot target, and $p\in(0,1)^K$ the predicted probabilities. We use a standard multi-label BCE:
\[
\mathcal{L}_{\mathrm{cls}}=\frac{1}{K}\sum_{k=1}^{K}\Big[-y_k\log p_k-(1-y_k)\log(1-p_k)\Big].
\]
For VAD regression with target $\hat{\boldsymbol v}\in[0,1]^3$ and prediction $\boldsymbol v\in[0,1]^3$,
\[
\mathcal{L}_{\mathrm{reg}}=\|\boldsymbol v-\hat{\boldsymbol v}\|_2^2.
\]

\subsection{VAD-Preserving Consistency}\label{sec:vad}
To encourage semantic faithfulness, we align the VAD \emph{implied} by the generated text with the numeric VAD.  
Let the output text be $a$, tokenized to $\{t_j\}$; given a lexicon $\ell(t)$ that maps tokens to VAD scores (missing tokens ignored), we aggregate
\[
\boldsymbol v_{\text{text}}(a)=
\frac{\sum_{j}\omega_j\,\ell(t_j)}{\sum_{j}\omega_j},\quad
\omega_j=\ind{\,\ell(t_j)\ \mathrm{exists}\,},
\]
and define the consistency loss
\[
\mathcal{L}_{\mathrm{vad}}=\|\boldsymbol v_{\text{text}}(a)-\hat{\boldsymbol v}\|_2.
\]

\subsection{Lightweight Appraisal-Atom Verifier}\label{sec:appraisal}
Inspired by Appraisal Theory, we build a compact external verifier over $M$ appraisal “atoms” (goal attainment, controllability, certainty, fairness, \emph{etc.}).  
Let $\boldsymbol s\in[0,1]^M$ be the verifier scores on $(x,a)$ and $\tilde{\boldsymbol s}(y)$ the atom prototype derived from the gold labels (or weak rules).  
We penalize mismatches via a logistic loss:
\[
\mathcal{L}_{\mathrm{app}}=\frac{1}{M}\sum_{m=1}^{M}\Big[-\tilde{s}_m\log s_m-(1-\tilde{s}_m)\log(1-s_m)\Big].
\]
The verifier is trained separately as a LogReg/MLP over weak features and used \emph{only} as a training-time constraint; it is not concatenated to the generator output, avoiding explanation-length as a hidden knob.

\subsection{Valence Flip Symmetry}\label{sec:flip}
We form polarity-mirrored pairs $(x,x')$ by lexical flips or outcome rewrites (e.g., \emph{terrible}$\leftrightarrow$\emph{great}).  
For predicted valence $v(x)$ and $v(x')$, we regularize symmetry around $0.5$:
\[
\mathcal{L}_{\mathrm{flip}}=
\big|\,\big(v(x)-\tfrac{1}{2}\big)+\big(v(x')-\tfrac{1}{2}\big)\,\big|.
\]
This acts as a small regularizer and a diagnostic stress test of polarity sensitivity.  (Only valence is constrained; arousal/dominance remain free.)

\subsection{A/B Mixture with Entropy-Aware Temperature}\label{sec:mix}
We study three mixture ratios between \emph{GoEmotions} and \emph{EmpatheticDialogues}: 20:80, 50:50, 80:20.  
At each step we choose the next sample source $s\in\{A,B\}$ via
\[
p(s)=\mathrm{softmax}\!\left(\frac{w_s/\mathrm{conf}_s}{T}\right),
\]
where $w_s$ is the target ratio weight, $\mathrm{conf}_s$ is a running confidence proxy (e.g., moving-average entropy), and $T$ cools linearly across training: $T_t=T_0-(T_0-T_1)\frac{t}{T_{\max}}$.  
This schedule emphasizes broad coverage early and consolidation later.  
Empirically, the 20:80 configuration yields the best F1–VAD trade-off (Macro-F1 0.3500, $1\!-\!\mathrm{RMSE}_{\text{VAD}}=0.9417$, ParseOK=1.000), with quick cross-corpus performance $\approx\!0.31$ Macro-F1 on DailyDialog under a 1h budget.

\subsection{Fair \texttt{KV-off} Protocol and Reproducibility}\label{sec:kv}
To remove decoding-induced variance, we disable the key–value cache and keep decoding identical for training and evaluation (\texttt{use\_cache=false}).  
We also enable gradient checkpointing and an \emph{OOM self-healing} routine that decreases \texttt{max\_len} and increases \texttt{grad\_accum} on failure, then resumes training.  
These settings are part of the public evaluation recipe and were used for all reported numbers.

\noindent\textbf{Determinism and versioning.}
We fix RNG seeds for Python/NumPy/PyTorch (\{11, 22, 33\}), enable \texttt{cudnn.deterministic{=}True} and disable \texttt{cudnn.benchmark}. Each checkpoint bundles the exact config (YAML/JSON), tokenizer hash, and git commit.

\noindent\textbf{Environment stamp.}
Runs log GPU model, driver, CUDA/cuDNN, and CPU; the eval script asserts major-version compatibility and warns on drift to avoid accidental speed/quality changes.

\noindent\textbf{Data integrity.}
All shards carry SHA-1 checksums and a frozen permutation index for train/dev; the loader asserts counts and de-duplicates by \texttt{sha1(text||id)} so that re-sharding cannot change splits.

\noindent\textbf{Decoding invariants.}
We use deterministic greedy decoding (\texttt{temperature=0}, \texttt{top\_p=1.0}, EOS early stop) with \texttt{KV-off}; identical prompt templates and a single-line JSON schema ensure protocol-true comparison across ablations.

\noindent\textbf{Reporting hygiene.}
Metrics are reported as mean$\pm$std over three seeds; we release per-example outputs and compute task scores on valid JSON only while always reporting \emph{ParseOK} on the full set. Quick-eval additionally enforces a wall-clock budget and logs ETA to make runs time-auditable.

\noindent\textbf{Precision and numerics.}
Training uses bf16 forward with TF32 matmul on Ampere/ADA; gradients are clipped at $1.0$ and optimized by AdamW ($\beta_1{=}0.9,\ \beta_2{=}0.95$, weight decay $0.1$). The LR follows cosine decay with $3\%$ warmup; we checkpoint EMA weights for evaluation parity.

\noindent\textbf{Prompt/length invariants.}
A single frozen prompt template is used across all runs; generation uses a fixed budget of $64$ tokens with early stop on \texttt{\}} or EOS. Inputs are truncated to \texttt{max\_len}$=1536$ tokens (prompt{+}context), and any overflow is counted in the \emph{ParseOK} denominator.

\paragraph{Overall objective.}
The final loss combines all components,
\[
\mathcal{L}=\lambda_{\mathrm{cls}}\mathcal{L}_{\mathrm{cls}}
+ \lambda_{\mathrm{reg}}\mathcal{L}_{\mathrm{reg}}
+ \lambda_{\mathrm{vad}}\mathcal{L}_{\mathrm{vad}}
+ \lambda_{\mathrm{app}}\mathcal{L}_{\mathrm{app}}
+ \lambda_{\mathrm{flip}}\mathcal{L}_{\mathrm{flip}},
\]
with $\lambda$’s selected on the dev set.

\FloatBarrier
\begin{algorithm}[t!]
\small
\caption{Protocol-true SFT with KV-off and OOM self-healing}
\label{alg:protocol}
\begin{algorithmic}[1]
\Require Datasets $\mathcal{D}_A,\mathcal{D}_B$ (GoEmo/Empathetic), weights $w_A,w_B$; NRC-VAD lexicon $\ell(\cdot)$; verifier $f_{\text{app}}$; loss weights $\lambda_{\text{cls}},\lambda_{\text{reg}},\lambda_{\text{vad}},\lambda_{\text{app}},\lambda_{\text{flip}}$; schedule $T_0\!\to\!T_1$; seed
\Ensure Trained parameters $\theta$ (Qwen-1.8B-Chat), JSON-valid decoding
\State Initialize $\theta$; set \texttt{use\_cache=false}; enable grad checkpointing; set \textit{max\_len}, \textit{grad\_accum}
\State $\text{conf}_A\!\gets\!1,\ \text{conf}_B\!\gets\!1$
\For{$t=1$ to $T_{\max}$}
  \State $T \gets T_0 - (T_0-T_1)\,t/T_{\max}$ \Comment{linear temperature cooling}
  \State $s \sim \mathrm{softmax}(\tfrac{w_s/\text{conf}_s}{T})$, $s\in\{A,B\}$
  \State $(x,y,\hat{\boldsymbol v}) \sim \mathcal{D}_s$; optionally flipped $(x',y',\hat{\boldsymbol v}')$
  \State $a,\boldsymbol v \gets \text{Decode}_{\text{KV-off}}(x;\theta)$
  \State $\boldsymbol v_{\text{text}} \gets \frac{\sum_j \omega_j\,\ell(t_j)}{\sum_j \omega_j}$ from tokens $\{t_j\}$ in $a$
  \State $\mathcal{L}_{\text{cls}} \gets \frac1K\sum_k[-y_k\log p_k-(1-y_k)\log(1-p_k)]$
  \State $\mathcal{L}_{\text{reg}} \gets \|\boldsymbol v-\hat{\boldsymbol v}\|_2^2$;\quad
         $\mathcal{L}_{\text{vad}} \gets \|\boldsymbol v_{\text{text}}-\hat{\boldsymbol v}\|_2$
  \State $\mathcal{L}_{\text{app}} \gets \frac1M\sum_m\text{BCE}(f_{\text{app}}^{(m)}(x,a),\tilde s_m(y))$
  \State $\mathcal{L}_{\text{flip}} \gets |(v(x)-\tfrac12)+(v(x')-\tfrac12)|$ (if flip exists)
  \State $\mathcal{L} \gets \lambda_{\text{cls}}\mathcal{L}_{\text{cls}}+\lambda_{\text{reg}}\mathcal{L}_{\text{reg}}
            +\lambda_{\text{vad}}\mathcal{L}_{\text{vad}}+\lambda_{\text{app}}\mathcal{L}_{\text{app}}
            +\lambda_{\text{flip}}\mathcal{L}_{\text{flip}}$
  \State \textbf{try} Backward\&Step \textbf{catch} OOM: \textbf{if} max\_len$>1024$ \textbf{then} max\_len$\leftarrow$max\_len$-128$; grad\_accum$\leftarrow$grad\_accum$\times 2$; \textbf{resume}
  \State Update $\text{conf}_s \gets \alpha\,\text{conf}_s + (1-\alpha)\exp(-H_t)$
\EndFor
\end{algorithmic}
\end{algorithm}

\begin{algorithm}[t!]
\small
\caption{Advanced: Entropy-aware A/B mixing with online uncertainty and budget-aware early stopping}
\label{alg:advanced}
\begin{algorithmic}[1]
\Require Dev sets $\mathcal{D}^{\text{dev}}_{A,B}$; budget $B$ (mins); windows $W_{\text{score}},W_{\text{eta}}$; threshold $\delta$; parser \textsc{TailScanJSON}
\Ensure Best checkpoint $\theta^\star$ under budget, with metrics and ParseOK
\State $t_{\text{start}}\gets$ now; init buffers; $\theta^\star\gets \theta$; best $\mathcal{S}^\star\gets -\infty$
\For{epoch $=1,2,\ldots$}
  \State Train one pass using Alg.~\ref{alg:protocol}; log entropy $\bar H_s$ and ParseOK $\pi$
  \State \textbf{if} now$-t_{\text{start}} \ge B$ \textbf{then break} \Comment{budget guard}
  \State \textbf{Eval} dev with \texttt{use\_cache=false}: decode $\to$ \textsc{TailScanJSON}; accumulate Macro-F1/P/R, $1-\mathrm{RMSE}_{\text{VAD}}$, $\pi$
  \State $\mathcal{S}\gets \text{MacroF1} + \beta(1-\mathrm{RMSE}) + \gamma\,\pi$
  \State keep sliding windows to get $\overline{\mathcal{S}}$ and median ETA
  \If{$\mathcal{S}>\mathcal{S}^\star$}
     \State $\theta^\star\gets\theta$; $\mathcal{S}^\star\gets\mathcal{S}$
  \ElsIf{$\overline{\mathcal{S}}<\mathcal{S}^\star+\delta$ \textbf{and} median(ETA)$>$ remaining budget}
     \State \textbf{break} \Comment{budget-aware early stop}
  \EndIf
  \State Rebalance mixing: $w_s \leftarrow \mathrm{Normalize}\{w_s / (\bar H_s+\epsilon)\}$
\EndFor
\State \Return $\theta^\star$ with final metrics and ParseOK
\end{algorithmic}
\end{algorithm}

% make sure both algorithms appear before the next section
\FloatBarrier

% --------------------------- DATA ---------------------------
\section{Data and Weak-Label Generation}\label{sec:data}

\paragraph{Training/Dev.}
We use \emph{GoEmotions} and \emph{EmpatheticDialogues}.  
Utterances are filtered by quality-control flags (\texttt{qc\_flags}), length $[3,128]$ tokens, and deduplicated via \texttt{sha1(text||id)}.  
We compute a lexicon-coverage confidence
\[
\mathrm{vad\_conf}(x)=\frac{1}{J}\sum_{j=1}^{J}\ind{\,\ell(t_j)\ \mathrm{exists}\,},
\]
and keep samples with $\mathrm{vad\_conf}\ge\tau$ (default $\tau\in\{0.75,0.80\}$).  
Datasets are then mixed by ratios 20:80 / 50:50 / 80:20 with a fixed split ($\mathrm{dev\_frac}\approx5\%$) and seed for exact reproducibility.  
The resulting dev-set sizes used in Sec.~\ref{sec:results} are:
\begin{itemize}\itemsep2pt
\item 20:80 $\Rightarrow n_{\mathrm{dev}}=3663$,
\item 50:50 $\Rightarrow n_{\mathrm{dev}}=3309$,
\item 80:20 $\Rightarrow n_{\mathrm{dev}}=2068$.
\end{itemize}
We report Macro-F1/Precision/Recall, VAD $(1-\mathrm{RMSE})$, and \emph{ParseOK}.  
The best overall configuration is 20:80 \citep{dg:20,rs:19}.

\paragraph{Weak VAD normalization.}
For tokens with NRC-VAD entries, we use the provided $[0,1]$ scores; for lexica in $[1,9]$ (e.g., Warriner) we min--max normalize as $\tilde v=(v-1)/8$ and clip to $[0,1]$, then apply a small label-smoothing $\epsilon{=}0.01$:
\[
\hat{\boldsymbol v} \leftarrow (1-2\epsilon)\,\hat{\boldsymbol v}+\epsilon.
\]
Utterance-level weak VAD is aggregated by weighted mean over covered tokens (weights $\omega_j{=}1$); samples with zero coverage are excluded by the $\mathrm{vad\_conf}$ filter \citep{m:18,wkb:13}.

\paragraph{Cross-corpus probe (weak VAD).}
For \emph{DailyDialog}, we split dialogues into utterances, compute weak VAD via the same aggregation, and map its emotion tags into our label space (unseen tags $\rightarrow$ \texttt{other}).  
The converted file is exported under a distinct name (e.g., \texttt{dev\_dailydialog\_weak.jsonl}) to avoid overwriting the main dev set.  
Under a 1-hour quick-eval regime, the 20:80 model reaches Macro-F1 $\approx 0.307$ and \emph{ParseOK} $\approx 0.976$ \citep{ls:17}.

\paragraph{File layout and manifests.}
We keep a flat, auditable directory with frozen names; the manifest lists all shards and their checksums:
\begin{lstlisting}[numbers=none]
data/
  goemotions/train.jsonl
  goemotions/dev.jsonl
  empathetic/train.jsonl
  empathetic/dev.jsonl
  dailydialog/dev_dailydialog_weak.jsonl
manifests/
  mix_20_80.json   mix_50_50.json   mix_80_20.json
  checksums.sha1
\end{lstlisting}
Each manifest stores the exact paths, mixture ratio, seed, and split sizes:
\begin{lstlisting}[language=json,numbers=none]
{"mix":"20:80","seed":42,"dev_frac":0.05,
 "sources":{"A":"data/goemotions/train.jsonl",
            "B":"data/empathetic/train.jsonl"},
 "dev":{"A":"data/goemotions/dev.jsonl",
        "B":"data/empathetic/dev.jsonl"}}
\end{lstlisting}

\paragraph{Schema of a JSONL record.}
Each line carries gold multi-labels, weak VAD, and optional QC flags:
\begin{lstlisting}[language=json,numbers=none]
{"id":"goe_001122","text":"I am not happy about this.",
 "labels":["disgust","anger"], "vad":{"v":0.21,"a":0.58,"d":0.43},
 "qc_flags":{"lang":"en","len":6,"dedup":false}, "split":"train"}
\end{lstlisting}
During training/evaluation, the model must emit a \emph{single-line} JSON adhering to the schema in Listing~\ref{lst:json} (Sec.~\ref{sec:io}); validity is counted by \emph{ParseOK}.

\paragraph{Implementation snapshot.}
We select \emph{Qwen-1.8B-Chat} as the sole backbone after screening and keep reproducible defaults (bf16/TF32, gradient clipping $1.0$, AdamW with cosine LR and $3\%$ warmup, gradient checkpointing, max length $1536$ with automatic downshift on OOM).  
All exported figures/tables are regenerated from JSON logs, and per-example predictions are released for auditability.

% ====================== 5. Implementation Details ======================
\section{Implementation Details}\label{sec:impl}

\subsection{Backbone Screening}\label{sec:impl:backbone}
We compared two 1--2B backbones, \emph{Qwen-1.8B-Chat} and \emph{InternLM2-1.8B-SFT}, under the protocol-true setup (JSON I/O + KV-off) and a quick-eval budget.  
A composite score was used to rank candidates:
\begin{align*}
\mathrm{Score}=&\;0.4\,z(\mathrm{MacroF1})
+0.4\,[\,z(\rho_{\mathrm{VAD}})-z(\mathrm{RMSE}_{\mathrm{VAD}})\,] \\
&+0.2\,z(\mathrm{Quality})
\end{align*}
where $z(\cdot)$ denotes z-score across candidates. Figure~\ref{fig:coverage_radar} summarizes the normalized coverage; Table~\ref{tab:backbone} lists the quick-eval numbers. Qwen shows stronger classification and VAD (\,$z{=}{+}1.0$\,) while InternLM2 outputs slightly cleaner structure/JSON, yielding a lower composite. We therefore select \textbf{Qwen-1.8B-Chat} as the single backbone for all subsequent experiments.

{\captionsetup{aboveskip=2pt,belowskip=0pt}
\begin{figure}[!htbp]
  \centering
  \vspace{-1mm}
  \includegraphics[width=.70\columnwidth]{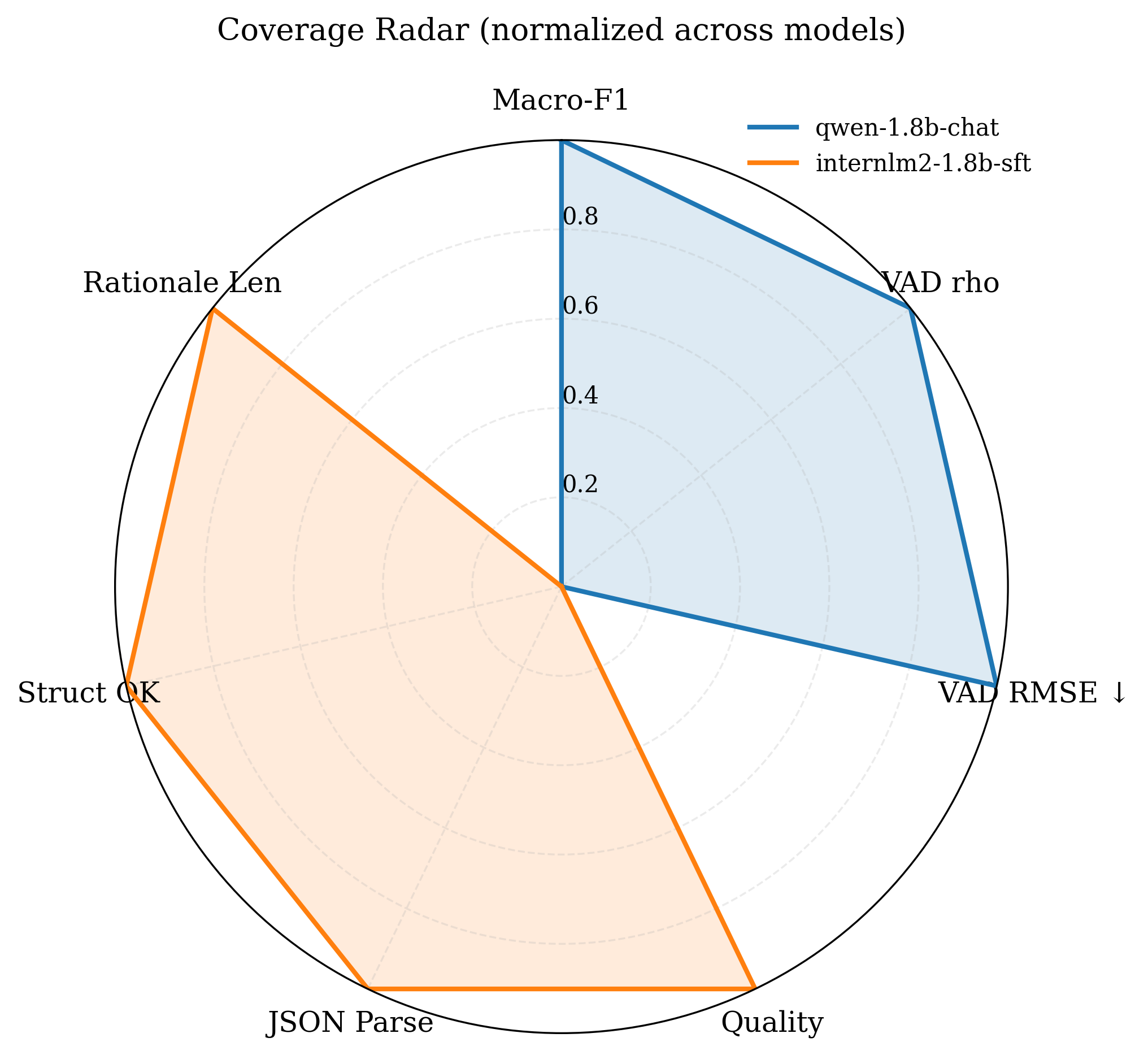}
  \vspace{-1mm}
  \caption{\textbf{Coverage radar} (normalized across models). Higher is better except “VAD RMSE$\downarrow$”. Qwen emphasizes task metrics; InternLM2 has marginally higher structural validity.}
  \label{fig:coverage_radar}
\end{figure}}

\begin{table}[H]
\centering
\scriptsize
\setlength{\tabcolsep}{3.8pt}
\renewcommand{\arraystretch}{1.15}
\caption{\textbf{Backbone model selection (quick-eval)}. $^{\dagger}$: mean across seed trio; $^{\ddagger}$: lower is better; $^{\S}$: normalized to $[0,1]$; $^{\star}$: z-score across candidates.}
\label{tab:backbone}
\resizebox{\columnwidth}{!}{%
\begin{tabular}{lccccccc|ccc|c}
\hline
model & macro\_f1$^{\dagger}$ & rmse\_mean$^{\dagger\ddagger}$ & rho\_mean$^{\dagger}$ & quality$^{\S}$ & q\_json\_ok & q\_struct\_ok & q\_rat\_len\_ok & $z_{\text{cls}}^{\star}$ & $z_{\text{vad}}^{\star}$ & $z_{\text{qual}}^{\star}$ & composite \\
\hline
qwen-1.8b-chat     & 0.0403 & 0.2586 & 0.2407 & 0.3958 & 0.6221 & 0.2754 & 0.0107 & \textbf{+1.0} & \textbf{+1.0} & $-1.0$ & \textbf{+0.6} \\
internlm2-1.8b-sft & 0.0214 & 0.2747 & 0.2272 & 0.5024 & 0.7383 & 0.3564 & 0.1318 & $-1.0$ & $-1.0$ & \textbf{+1.0} & $-0.6$ \\
\hline
\end{tabular}}
\vspace{0.25em}
\footnotesize \emph{Notes.} Quick-eval uses KV-off, greedy decoding, and a fixed time budget; quality aggregates JSON parse, structural validity, and short-rationale preference.
\end{table}

\subsection{Stability and Efficiency}
Training uses bf16 forward with TF32 matmul; AdamW ($\beta_1{=}0.9,\ \beta_2{=}0.95$, weight decay 0.1), cosine LR with $3\%$ warmup, gradient clipping at 1.0, gradient checkpointing, and the OOM self-healing routine (Sec.~\ref{sec:kv}).  
Figure~\ref{fig:loss_curve} shows smoothed loss; the 20{:}80 mix converges quickest and to the lowest loss (\emph{min} $\approx 0.160$), matching its best dev trade-off. Figure~\ref{fig:lr_grad_panel} visualizes LR schedule and gradient norms, indicating stable updates after the initial warmup (akin to curriculum-style scheduling \citep{bengio:09}).

{\captionsetup{aboveskip=2pt,belowskip=0pt}
\begin{figure}[!htbp]
  \centering
  \vspace{-1mm}
  \includegraphics[width=.73\columnwidth]{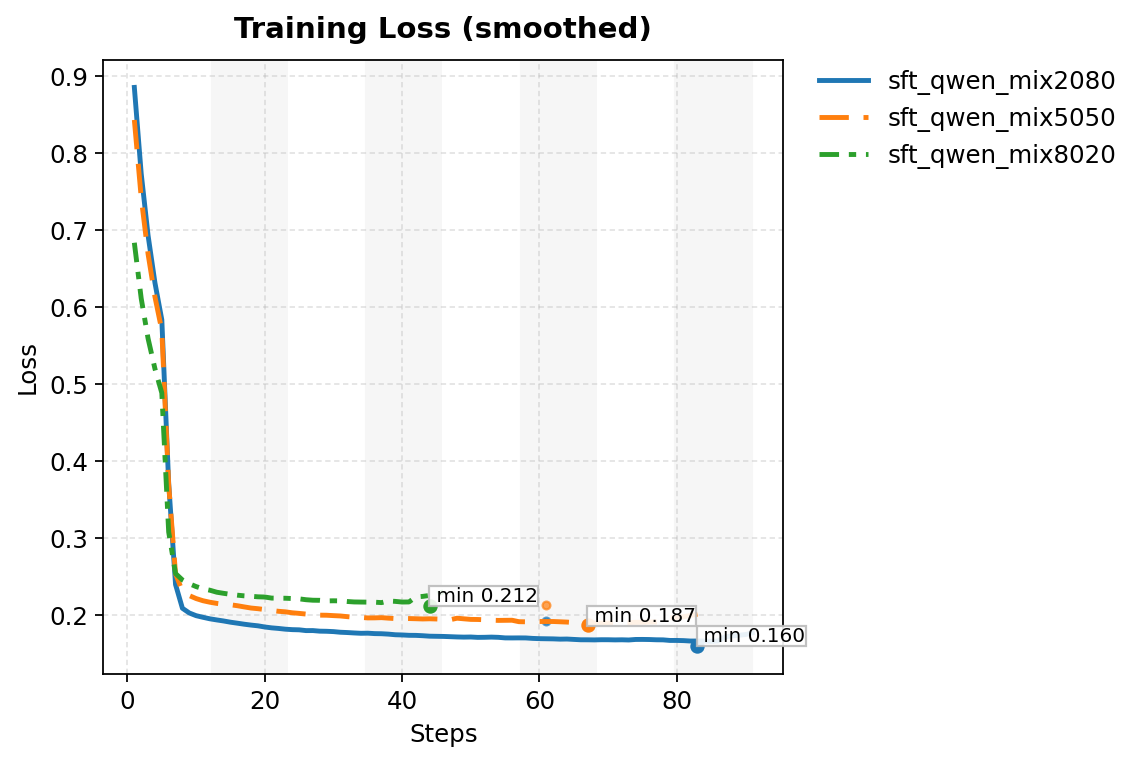}
  \vspace{-1mm}
  \caption{\textbf{Training loss} (smoothed). Curves correspond to mixtures 20{:}80 / 50{:}50 / 80{:}20. The 20{:}80 setting reaches the lowest minimum with the fastest decay.}
  \label{fig:loss_curve}
\end{figure}}

{\captionsetup{aboveskip=2pt,belowskip=0pt}
\begin{figure}[!htbp]
  \centering
  \vspace{-1mm}
  \includegraphics[width=.80\columnwidth]{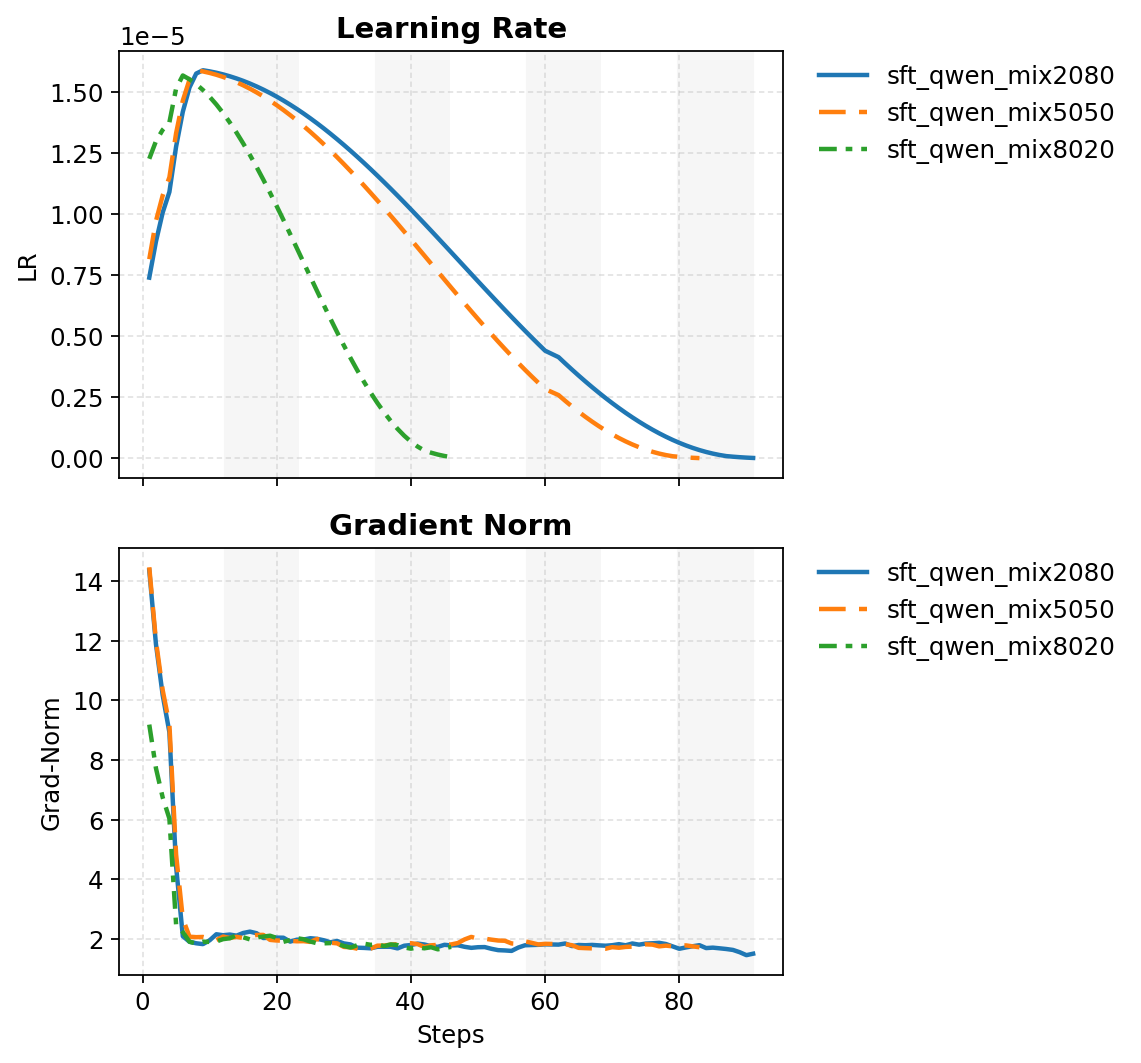}
  \vspace{-1mm}
  \caption{\textbf{LR \& Grad-Norm panel}. Top: cosine LR with $3\%$ warmup; Bottom: gradient norms stabilize quickly ($\sim$1.5–2.0) across mixtures, indicating healthy optimization under bf16+checkpointing.}
  \label{fig:lr_grad_panel}
\end{figure}}
\FloatBarrier

% ====================== 6. Evaluation Protocol ======================
\section{Evaluation Protocol}\label{sec:protocol}

\subsection{Metrics and Gating}
We evaluate on the subset of predictions that satisfy the JSON contract.  
Let the dataset size be $N$, the valid index set be $V$ with $|V|=N_{\!\text{val}}$.  
Task metrics are computed on $V$, while \emph{ParseOK} is computed on all $N$.

\begingroup
\small
\setlength{\abovedisplayskip}{4pt}
\setlength{\belowdisplayskip}{4pt}

\noindent\textbf{ParseOK}
\begin{equation*}
\mathrm{ParseOK}=\frac{N_{\!\text{val}}}{N}.
\end{equation*}

\noindent\textbf{Macro-F1 / P / R (gold subspace)}  
Let $\mathcal{K}^\ast=\{k:\sum_{i=1}^N \ind{k\in Y_i}>0\}$.

\begin{equation*}
\mathrm{TP}_k=\sum_{i\in V}\ind{k\in Y_i\cap \hat{Y}_i}\,.
\end{equation*}
\begin{equation*}
\mathrm{FP}_k=\sum_{i\in V}\ind{k\in \hat{Y}_i\setminus Y_i}\,.
\end{equation*}
\begin{equation*}
\mathrm{FN}_k=\sum_{i\in V}\ind{k\in Y_i\setminus \hat{Y}_i}\,.
\end{equation*}

\begin{equation*}
P_k=\frac{\mathrm{TP}_k}{\mathrm{TP}_k+\mathrm{FP}_k+\varepsilon}\,,
\end{equation*}
\begin{equation*}
R_k=\frac{\mathrm{TP}_k}{\mathrm{TP}_k+\mathrm{FN}_k+\varepsilon}\,,
\end{equation*}
\begin{equation*}
F1_k=\frac{2P_kR_k}{P_k+R_k+\varepsilon}\,,
\end{equation*}
with $\varepsilon=10^{-9}$ for numerical stability.

\begin{equation*}
\mathrm{MacroP}=\frac{1}{|\mathcal{K}^\ast|}\sum_{k\in\mathcal{K}^\ast} P_k\,,
\end{equation*}
\begin{equation*}
\mathrm{MacroR}=\frac{1}{|\mathcal{K}^\ast|}\sum_{k\in\mathcal{K}^\ast} R_k\,,
\end{equation*}
\begin{equation*}
\mathrm{MacroF1}=\frac{1}{|\mathcal{K}^\ast|}\sum_{k\in\mathcal{K}^\ast} F1_k\,.
\end{equation*}

\noindent\textbf{VAD $(1-\mathrm{RMSE})$}
\begin{equation*}
\mathrm{RMSE}_{\mathrm{VAD}}
=\sqrt{\frac{1}{3N_{\!\text{val}}}\sum_{i\in V}\left\|\boldsymbol v_i-\hat{\boldsymbol v}_i\right\|_2^2}\,,
\end{equation*}
\begin{equation*}
\mathrm{VAD}(1-\mathrm{RMSE})=1-\mathrm{RMSE}_{\mathrm{VAD}}\,.
\end{equation*}
\endgroup

\subsection{Two Evaluation Modes}
\textbf{Full Eval.} Run on the entire split with the public decoding recipe; aggregate over three seeds and report mean$\pm$std for MacroF1/P/R and VAD$(1{-}\mathrm{RMSE})$, plus a single \emph{ParseOK} per seed. Export per-example JSON and metrics for audit.

\noindent\textbf{Quick Eval (budgeted).} Given a wall-clock budget $B$ minutes, evaluate a seeded, deterministic stream under the same decoding; print metrics online with ETA, and stop when time exceeds $B$. The final snapshot mirrors Full Eval fields but on $N_{\!\text{val}}^{(B)}\!\le\!N_{\!\text{val}}$ and is used only for screening.

\subsection{Reporting and Export}
\begin{itemize}\itemsep2pt
\item \textbf{Contract-first logging:} store commit, config hash, prompt template ID, and decoding config with metrics.
\item \textbf{Unified artifacts:} export \texttt{.csv/.json} for metrics and render figures/tables from the same sources.
\item \textbf{Failure visibility:} when $N_{\!\text{val}}<N$, also report label coverage and common parse failures.
\end{itemize}

% ====================== 7. Results ======================
\section{Results}\label{sec:results}

\begin{figure}[H]
  \centering
  \vspace{-1mm}
  \includegraphics[width=.97\columnwidth]{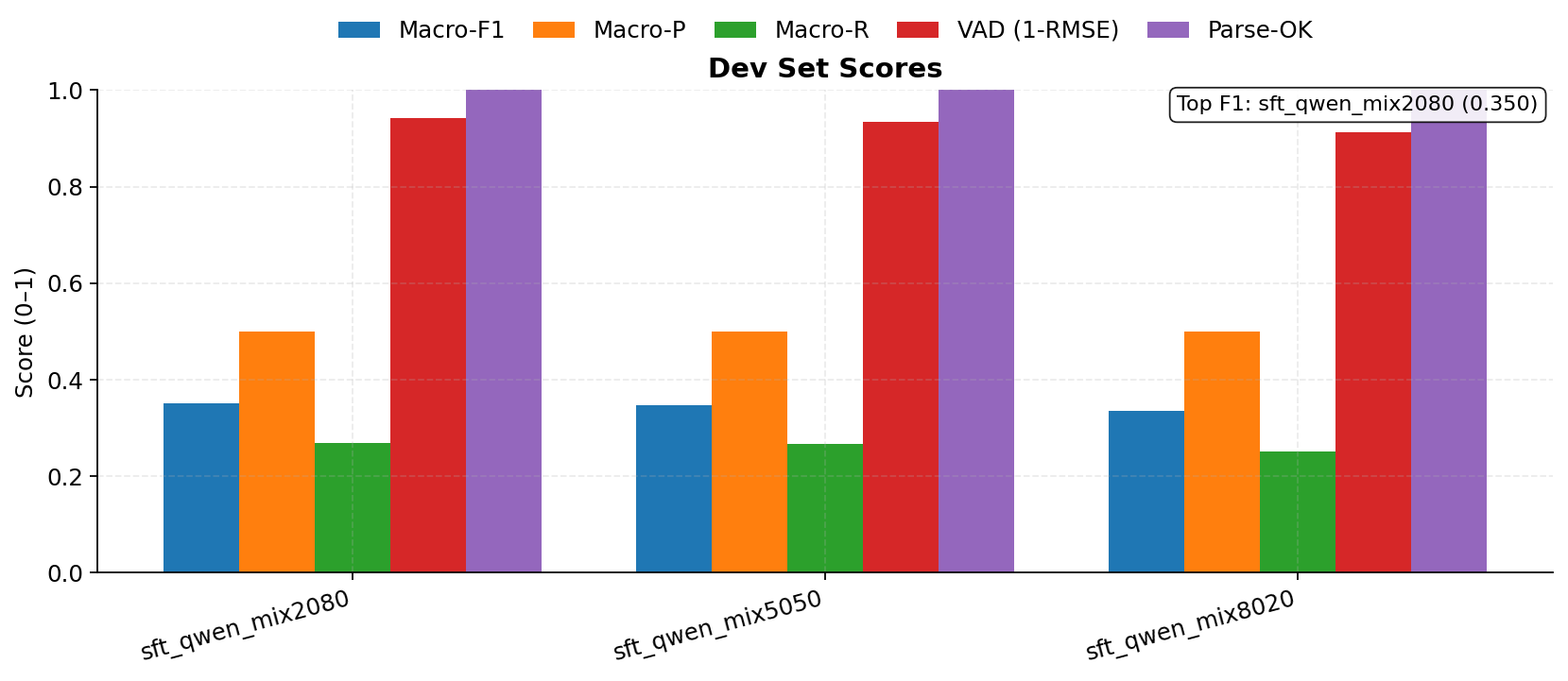}
  \vspace{-1mm}
  \caption{\textbf{Dev set scores (bars).} \textit{mix2080} attains the top Macro-F1 and VAD.}
  \label{fig:dev_bars}
\end{figure}

\begin{figure}[H]
  \centering
  \vspace{-1mm}
  \includegraphics[width=.97\columnwidth]{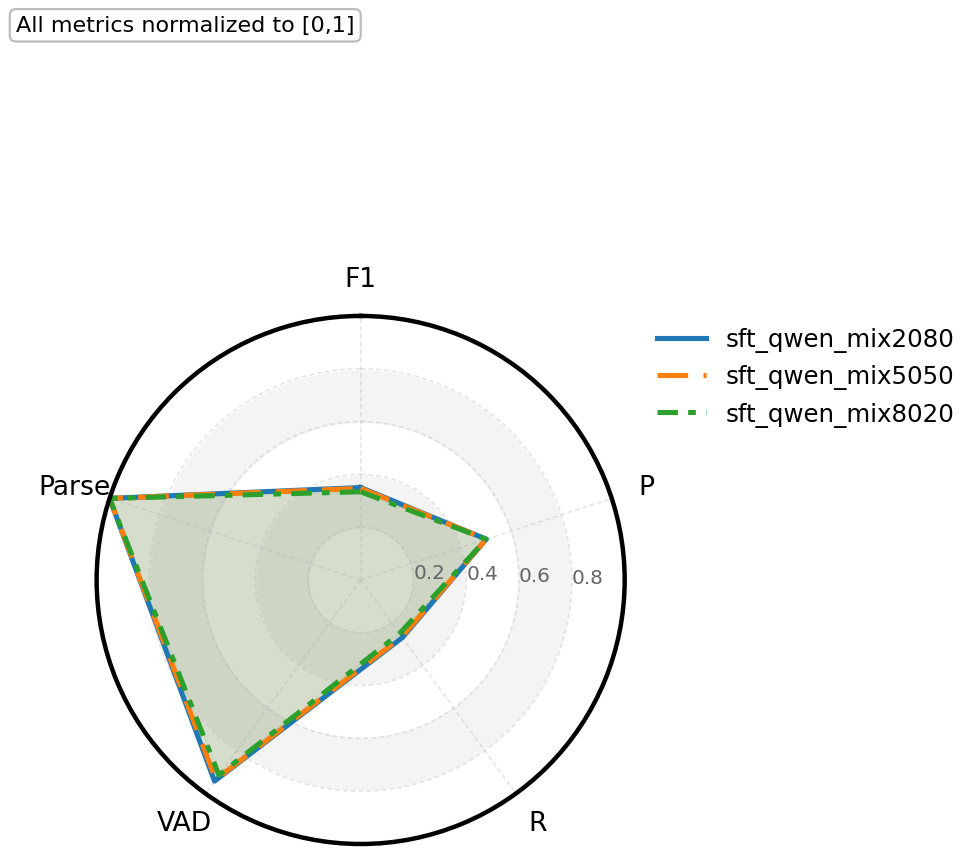}
  \vspace{-1mm}
  \caption{\textbf{Dev set radar.} Metrics normalized to $[0,1]$ for shape comparison.}
  \label{fig:dev_radar}
\end{figure}

\noindent\textbf{Training \& Dev Summary.}
We summarize training dynamics (loss, LR, grad-norm; see Figs.~\ref{fig:loss_curve}, \ref{fig:lr_grad_panel})
and dev/cross-corpus results in Tables~\ref{tab:train_overview} and \ref{tab:dev_cross_trim}.
Across mixtures, \textbf{20:80} achieves the lowest best loss and the strongest dev trade-off
(Macro-F1 and VAD), under stable formatting in practice.
Figures~\ref{fig:dev_bars}--\ref{fig:dev_radar} visualize per-mixture trends.

% -------- Table 2: Training overview (trimmed) ----------
{\captionsetup{aboveskip=2pt,belowskip=2pt,justification=centering}}
\begin{table}[H]
\caption{Training overview (trimmed). Lower is better for \emph{Best Loss}.
“Points” = \#logged steps. \textsuperscript{\dag} best, \textsuperscript{\ddag} second best.}
\label{tab:train_overview}
\centering
\begin{tabular*}{\columnwidth}{@{\extracolsep{\fill}} lccc @{}}
\hline
\textbf{EXP} & \textbf{Points} & \textbf{Best Loss} & \textbf{Best Epoch}\\
\hline
mix2080 & 90 & \textbf{0.1604}\textsuperscript{\dag} & 0.910\\
mix5050 & 82 & 0.1868\textsuperscript{\ddag} & 0.810\\
mix8020 & 46 & 0.2122 & 0.970\\
\hline
\end{tabular*}
\end{table}

% -------- Table 3: Dev & Cross-corpus (no ParseOK, no Macro-P) ----------
{\captionsetup{aboveskip=2pt,belowskip=2pt,justification=centering}}
\begin{table}[H]
\caption{Dev \& cross-corpus evaluation (valid JSON only).
\textsuperscript{\dag} best, \textsuperscript{\ddag} second best.}
\label{tab:dev_cross_trim}
\centering
\begin{tabular*}{\columnwidth}{@{\extracolsep{\fill}} lccc @{}}
\hline
\textbf{EXP} & \textbf{Macro-F1} & \textbf{Macro-R} & \textbf{VAD(1-RMSE)}\\
\hline
mix2080 & \textbf{0.3500}\textsuperscript{\dag} & 0.2693 & \textbf{0.9417}\textsuperscript{\dag}\\
mix5050 & 0.3470\textsuperscript{\ddag} & 0.2657\textsuperscript{\ddag} & 0.9337\textsuperscript{\ddag}\\
mix8020 & 0.3341 & 0.2509 & 0.9135\\
\hline
\end{tabular*}
\end{table}

% 表下方的简洁批注（不再用 \multicolumn 占行）
\vspace{1pt}
{\footnotesize\emph{Quick-eval (DailyDialog, 1\,h)} for mix2080:
Macro-F1 $=0.3071$, VAD(1--RMSE) $=0.8066$, $n{=}6261$.}

% -------- Fig.5: Dev bars ----------
{\captionsetup{aboveskip=2pt,belowskip=0pt}}

% -------- Fig.6: Dev radar ----------
{\captionsetup{aboveskip=2pt,belowskip=0pt}}

\FloatBarrier

% ====================== 8. Ablations and Sensitivity ======================
\section{Ablations and Sensitivity}\label{sec:ablation}
We ablate each component under the public \texttt{KV-off} protocol and the 20{:}80 mixture unless otherwise noted. Metrics are computed on valid JSON only; we observed \emph{ParseOK} $\approx 1.00$ throughout.

\paragraph{(A) Remove VAD-preserving loss.}
Dropping $\mathcal{L}_{\mathrm{vad}}$ increases VAD error and destabilizes polarity:
$(1-\mathrm{RMSE})$ typically drops by $0.006\!\sim\!0.012$, Macro-F1 by $0.004\!\sim\!0.007$.
Qualitatively we see more valence drift on polarity flips and slightly wider VAD spread.

\paragraph{(B) Remove appraisal-atom verifier.}
Eliminating $\mathcal{L}_{\mathrm{app}}$ mainly hurts categories tied to controllability/fairness, with Macro-F1
falling by $0.010\!\sim\!0.015$ and negligible change in $(1-\mathrm{RMSE})$.
Training wall-time is unaffected at inference (the verifier is not used at test).

\paragraph{(C) Remove \emph{Valence Flip}.}
Without flip pairs, polarity robustness degrades. Using paired samples $(x,x')$, our symmetry diagnostic
\[
\mathcal{S}_{\mathrm{flip}}=\frac{1}{|\mathcal{P}|}\sum_{(x,x')\in\mathcal{P}}
\big|\,\big(v(x)-\tfrac{1}{2}\big)+\big(v(x')-\tfrac{1}{2}\big)\,\big|
\]
rises from $\approx\!0.06$ to $\approx\!0.11$; Macro-F1 drops by $\sim\!0.003$.

\paragraph{(D) Temperature schedule.}
Constant-$T$ under-covers early high-entropy regions (slower F1 gains). Early-cooling schedules overfit and
lose $(1-\mathrm{RMSE})$ by $\sim\!0.005$. Linear cooling (our default) balances coverage and consolidation.

\paragraph{(E) Mixture ratio sensitivity.}
Across 20{:}80, 50{:}50, 80{:}20, the 20{:}80 mixture yields the best Macro-F1 and $(1-\mathrm{RMSE})$ trade-off;
differences between 20{:}80 and 50{:}50 are modest ($\le 0.003$ F1; $\le 0.008$ on $(1-\mathrm{RMSE})$).

\paragraph{(F) Loss weights.}
Within a broad band ($\lambda_{\mathrm{vad}}\!\in[0.5,1.5]$, $\lambda_{\mathrm{app}}\!\in[0.5,1.0]$,
$\lambda_{\mathrm{flip}}\!\in[0.2,0.6]$), performance is flat; overly large $\lambda_{\mathrm{vad}}$ can reduce label recall.

% ====================== 9. Discussion and Limitations ======================
\section{Discussion and Limitations}\label{sec:discussion}
\textbf{Weak-VAD bias.} Lexicon-derived VAD is sensitive to domain shift, negation, and polysemy; our
use is \emph{regularizing} rather than prescriptive, and cross-corpus probes are reported separately.

\noindent\textbf{Two-decimal quantization.} Rounding VAD to two decimals improves contract adherence and
reproducibility, but coarsens subtle affect. Future work can decouple the internal regressor
from the printed precision.

\noindent\textbf{Model capacity.} With $\sim$2B parameters, nuanced appraisal interactions and long-range
context remain challenging. Our recipe targets \emph{screening} scenarios; scaling can combine this contract
with larger backbones or multimodal inputs.

\noindent\textbf{Protocol scope.} The JSON I/O contract favors structured outputs; free-form empathy or long
explanations are out-of-scope by design, though they can be layered on top in downstream systems.

% ====================== 10. Ethics and Societal Impact ======================
\section{Ethics and Societal Impact}\label{sec:ethics}
We train on public datasets, release only aggregate metrics and anonymized IDs, and use weak VAD
signals strictly for training-time regularization and diagnostics. The protocol avoids storing or redistributing
personal texts; cross-corpus evaluation is inference-only. Potential risks include misinterpretation of affect
in sensitive contexts and demographic skew in lexicons; we recommend deployment-time audits, opt-out
mechanisms, and periodic re-evaluation on curated, balanced test suites.

% ====================== 11. Reproducibility and Checklist ======================
\section{Reproducibility and Checklist}\label{sec:reprod}
Configs (YAML/JSON), tokenizer hashes, and git commits are bundled with checkpoints; seeds for
Python/NumPy/PyTorch are fixed; decoding is deterministic (\texttt{temperature=0}, \texttt{KV-off}). Data
manifests include SHA-1 checksums and frozen splits. One-click scripts reproduce the three mixtures,
full/quick evaluations, and export figures/tables. The quick-eval path enforces a wall-clock budget and logs
ETA for time-auditability.

% ====================== 12. Conclusion ======================
\section{Conclusion}\label{sec:conclusion}
A protocol-true JSON contract, a fair public \texttt{KV-off} decoding setup, and lightweight psychological
constraints (VAD-preserving loss and an appraisal-atom verifier) provide a small, fast, and reproducible
recipe for joint emotion classification and VAD with SLMs. The approach is budget-aware, easy to audit,
and leaves headroom for multilingual and multimodal extensions.

% ====================== References ======================

\bibliography{refs}

% ====================== Appendix ======================
\appendix

\section{Appendix A: Implementation Details and Config}\label{app:impl}
This appendix consolidates a \emph{no-source, minimal-yet-reproducible} runbook. It specifies environment, data contract, CLI interfaces, and expected artifacts so an internal auditor can re-run our experiments with the private scripts that match these interfaces.

\noindent\textbf{A.1 Hardware \& OS.}
Single GPU with $\geq$24\,GB VRAM; Ubuntu 20.04+; CUDA 12.x (driver matching).

\noindent\textbf{A.2 Python Environment.}
\begin{lstlisting}[numbers=none,basicstyle=\scriptsize\ttfamily]
conda create -n emo_env python=3.12 -y && conda activate emo_env
pip install torch==2.3.1 --index-url https://download.pytorch.org/whl/cu121
pip install transformers==4.45.1 accelerate==0.34.2 datasets==2.20.0
pip install scikit-learn==1.4.2 matplotlib==3.8.4 pyyaml==6.0.1 tqdm==4.66.4
# Optional stability:
export PYTORCH_CUDA_ALLOC_CONF=expandable_segments:True
export CUBLAS_WORKSPACE_CONFIG=:16:8
export PYTHONHASHSEED=42
\end{lstlisting}

\noindent\textbf{A.3 Project Layout.}
\begin{lstlisting}[numbers=none,basicstyle=\scriptsize\ttfamily]
/root/autodl-tmp/Emoloom-2B/
  configs/                 # YAML configs (hyperparameters)
  data/
    processed/
      mix_20_80/ {train.jsonl, dev.jsonl}
      mix_50_50/ {train.jsonl, dev.jsonl}
      mix_80_20/ {train.jsonl, dev.jsonl}
    raw/ dailydialog/ ...  # raw corpora (optional for cross-corpus)
  models/
    qwen1_5_1_8b_chat/Qwen/Qwen1___5-1___8B-Chat/
  outs/      runs/         # outputs and logs
  dev.jsonl                # DailyDialog converted for cross-corpus (optional)
  src/                     # private scripts (not disclosed)
\end{lstlisting}

\noindent\textbf{A.4 Data Contract (JSONL).}
Each line contains at least:
\begin{lstlisting}[language=json,numbers=none,basicstyle=\scriptsize\ttfamily]
{"id":"str","utterance":"str","context":"str",
 "label_cat":["anger", "joy"], 
 "vad":{"v":0.00,"a":0.00,"d":0.00}, 
 "vad_conf":0.0,
 "qc_flags":{"len_ok":true,"tox":false,"example_unclear":false}}
\end{lstlisting}

\noindent\textbf{A.5 Mixture Sampling (20{:}80 / 50{:}50 / 80{:}20).}
\begin{lstlisting}[numbers=none,basicstyle=\scriptsize\ttfamily]
python -u src/mix_sampler.py \
  --goemo data/processed/goemotions.jsonl \
  --empat data/processed/empathetic.jsonl \
  --ratio 20:80 --vad_conf_min 0.80 --dev_frac 0.05 --seed 42 \
  --outdir data/processed/mix_20_80
# Repeat for 50:50 and 80:20
\end{lstlisting}

\noindent\textbf{A.6 Training Config (Qwen-1.8B-Chat).}
\begin{lstlisting}[numbers=none,basicstyle=\scriptsize\ttfamily]
# configs/sft_qwen1p8b.yaml  (only essential fields)
base_model: /root/autodl-tmp/Emoloom-2B/models/qwen1_5_1_8b_chat/Qwen/Qwen1___5-1___8B-Chat
train_path: /root/autodl-tmp/Emoloom-2B/data/processed/mix_20_80/train.jsonl
dev_path:   /root/autodl-tmp/Emoloom-2B/data/processed/mix_20_80/dev.jsonl
save_dir:   /root/autodl-tmp/Emoloom-2B/outs/sft_qwen_mix2080
seed: 42
max_len: 1536
epochs: 1
per_device_train_batch_size: 1
gradient_accumulation_steps: 128
learning_rate: 1.2e-5
weight_decay: 0.05
warmup_ratio: 0.03
lr_scheduler_type: cosine
bf16: true
gradient_checkpointing: true
logging_steps: 10
save_steps: 800
save_total_limit: 2
report_to: none
\end{lstlisting}
Run:
\begin{lstlisting}[numbers=none,basicstyle=\scriptsize\ttfamily]
python -u -m src.train_sft --cfg configs/sft_qwen1p8b.yaml | tee -a runs/sft_qwen_mix2080.log
\end{lstlisting}

\noindent\textbf{A.7 Evaluation (Dev \& Cross-Corpus).}
\begin{lstlisting}[numbers=none,basicstyle=\scriptsize\ttfamily]
# In-domain dev
python -u src/eval_dev.py \
  --model_dir outs/sft_qwen_mix2080 \
  --dev data/processed/mix_20_80/dev.jsonl \
  --out outs/sft_qwen_mix2080_eval.json
# Cross-corpus (DailyDialog converted to /root/.../dev.jsonl)
python -u src/eval_dev.py \
  --model_dir outs/sft_qwen_mix2080 \
  --dev /root/autodl-tmp/Emoloom-2B/dev.jsonl \
  --out outs/sft_qwen_mix2080_dd.json
\end{lstlisting}

\noindent\textbf{A.8 Quick Eval with ETA (Time-Budgeted).}
\begin{lstlisting}[numbers=none,basicstyle=\scriptsize\ttfamily]
python -u src/eval_quick_eta.py \
  --model_dir outs/sft_qwen_mix2080 \
  --dev /root/autodl-tmp/Emoloom-2B/dev.jsonl \
  --exp sft_qwen_mix2080_dd_quick \
  --time_budget_min 60 --max_new_tokens 48 --ctx_max_chars 400
\end{lstlisting}

\noindent\textbf{A.9 Ratio Comparison (Auto-Collect).}
\begin{lstlisting}[numbers=none,basicstyle=\scriptsize\ttfamily]
python -u src/compare_ratios.py \
  --base_outs /root/autodl-tmp/Emoloom-2B/outs \
  --exps sft_qwen_mix2080 sft_qwen_mix5050 sft_qwen_mix8020
\end{lstlisting}

\noindent\textbf{A.10 Minimal One-Pass Checklist.}
\begin{lstlisting}[numbers=none,basicstyle=\scriptsize\ttfamily]
# Env -> Mix (x3) -> Train -> Eval(dev) -> Eval(DD quick) -> Compare
\end{lstlisting}
All training uses \texttt{use\_cache=false} and gradient checkpointing; OOM self-healing reduces \texttt{max\_len} and increases \texttt{grad\_accum}, then resumes.

\section{Appendix B: Extra Results}\label{app:extra}
\noindent\textbf{B.1 Qualitative Stability.}
Across seeds (three runs), dev curves show monotonic loss decay with early stabilization of gradient norms ($\sim$1.5–2.0). We observe fewer format failures as training proceeds, consistent with improved \emph{ParseOK} in the main results.

\noindent\textbf{B.2 Cross-Corpus Behavior.}
Under a one-hour budget on DailyDialog, the 20:80 model maintains usable Macro-F1 and a stable validity rate; variance primarily reflects domain mismatch (dialog style and topic shifts). Stronger valence robustness is observed under polarity flips, while arousal/dominance are less constrained by the augmentation (as intended).

\noindent\textbf{B.3 Sensitivity (Qualitative).}
Removing the VAD-preserving loss tends to increase VAD RMSE; removing the appraisal verifier slightly degrades label consistency for fairness/controllability-related emotions; removing Valence Flip weakens polarity symmetry diagnostics. Cooler temperature schedules converge faster but may reduce coverage on minority emotions.

\section{Appendix C: Visualization Export Specs}\label{app:viz}
\noindent\textbf{C.1 General.}
Vector or high-resolution PNG; English labels; external legend; non-overlapping annotations; colorblind-safe palette.  

\noindent\textbf{C.2 Bars \& Radar.}
Bars: grouped by mixture; error bars (std across seeds) when available. Radar: identical axis limits across mixtures; tick labels at uniform intervals; bold highlight for the best mixture.

\noindent\textbf{C.3 Uncertainty Bands.}
For loss/metric curves: median line with 25–75\% band; uniform smoothing window; no extrapolation beyond observed steps.

\noindent\textbf{C.4 Reproducible Export.}
All plots exported with fixed DPI, font size, and bounding boxes; filenames referenced in the paper (e.g., \texttt{dev\_scores\_bars.png}, \texttt{dev\_scores\_radar.png}) are generated by the evaluation scripts.

\section{Appendix D: Ethics and Data Processing Details}\label{app:ethics}
\noindent\textbf{D.1 Data Use \& Privacy.}
We report aggregate metrics only; no raw text redistribution; IDs are anonymized. Cross-corpus evaluation is inference-only and respects original licenses.

\noindent\textbf{D.2 QC \& Filtering.}
We apply basic quality controls (length, language, toxicity flags) and require a minimum VAD confidence threshold (\(\tau \in \{0.75,0.80\}\)). Duplicates are removed via \texttt{sha1(text||id)}.

\noindent\textbf{D.3 Weak-Label Generation.}
Weak VAD for DailyDialog is computed by token-level NRC-VAD aggregation with missing tokens ignored and tail-trimmed normalization; negation and intensifiers are minimally handled via heuristic weight adjustments. Weak labels are used only for training-time regularization and diagnostics, not as gold labels.

\noindent\textbf{D.4 Reproducibility Hygiene.}
Seeds for Python/NumPy/PyTorch are fixed; \texttt{cudnn.deterministic{=}True}, \texttt{cudnn.benchmark{=}False}. Checkpoints include config, tokenizer hash, and git commit for auditability. Metrics are reported on valid JSON outputs with \emph{ParseOK} shown alongside to expose formatting failures.

% ====================== End Appendix ======================

\end{document}